%% file: acl_latex.tex
\pdfoutput=1

\documentclass[11pt]{article}

\usepackage[final]{acl}

\usepackage{times}
\usepackage{latexsym}

\usepackage[T1]{fontenc}

\usepackage[utf8]{inputenc}

\usepackage{microtype}

\usepackage{inconsolata}

\usepackage{graphicx}
\usepackage{multirow}
\usepackage{tcolorbox}
\usepackage{amsmath}
\usepackage{xcolor}
\usepackage{pifont}
\definecolor{powderblue}{RGB}{220, 230, 241}
\definecolor{PaleSageGreen}{HTML}{D8E2C6}
\definecolor{MutedApricot}{RGB}{243,217,193}

\newcommand{\myparagraph}[1]{\textbf{#1}\hspace{0.4em}}

\usepackage{booktabs}

\title{Vision-and-Language Navigation with Analogical \\ Textual Descriptions in LLMs}

\author{
Yue Zhang\textsuperscript{1} \quad Tianyi Ma\textsuperscript{1} \quad Zun Wang\textsuperscript{2} \quad
\textbf{Yanyuan Qiao\textsuperscript{3}} \quad 
\textbf{Parisa Kordjamshidi\textsuperscript{1}} \\
\textsuperscript{1}Michigan State University \quad
\textsuperscript{2}UNC Chapel Hill\quad
\textsuperscript{3} University of Adelaide \\
\texttt{zhan1624@msu.edu}
}

\begin{document}
\maketitle

\input{sections/abstract}
\input{sections/intro}
\input{sections/related_work}
\input{sections/method}
\input{sections/experiment}
\input{sections/conclusion}

\bibliography{custom}
\clearpage

\appendix

\input{sections/appendix}

\end{document}

%% file: sections/abstract.tex
\begin{abstract}
Integrating large language models (LLMs) into embodied AI models is becoming increasingly prevalent. However, existing zero-shot LLM-based Vision-and-Language Navigation (VLN) agents either encode images as textual scene descriptions, potentially oversimplifying visual details, or process raw image inputs, which can fail to capture  abstract semantics required for high-level reasoning. 
In this paper, we improve the navigation agent’s contextual understanding by incorporating textual descriptions from multiple perspectives that facilitate analogical reasoning across images.  By leveraging text-based analogical reasoning, the agent enhances its global scene understanding and spatial reasoning, leading to more accurate action decisions. We evaluate our approach on the R2R dataset, where our experiments demonstrate significant improvements in navigation performance.

\end{abstract}

%% file: sections/intro.tex
\section{Introduction}
With the LLMs being applied across diverse domains~\cite{yu2024crema, zhangspartun3d,zhang2024common, guo2025rethinking}, their integration into VLN agents has emerged as a promising development. Zero-shot LLM-based VLN agents represent a significant shift from traditional navigation agents that rely on extensive task-specific training, demonstrating greater adaptability and generalizability to a wide range of environments~\cite{zhang2024visionandlanguage}. 

\begin{figure}
    \centering
\includegraphics[width=\linewidth]{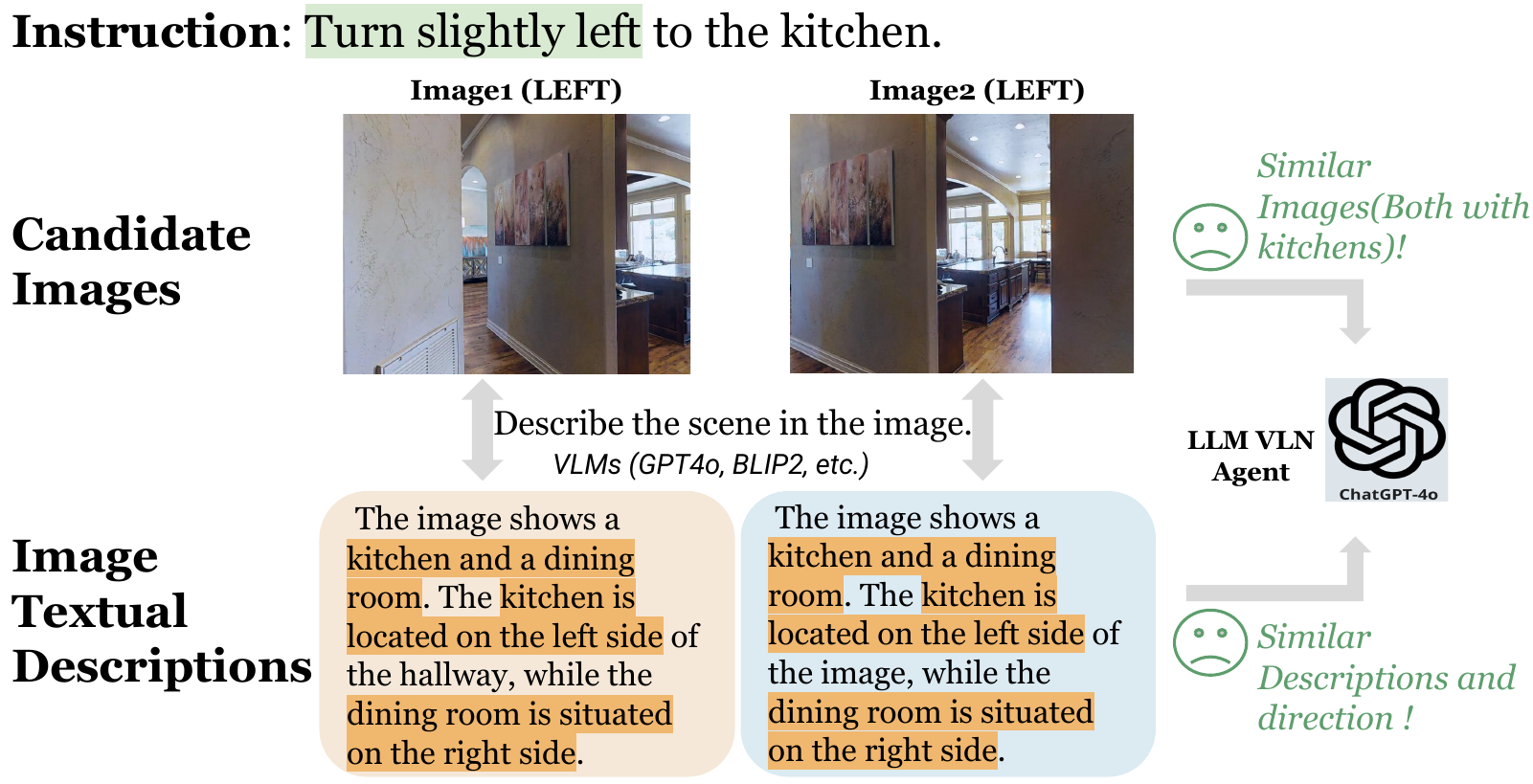}
    \caption{Challenges in current LLM-based VLN Agent. The highlighted orange text shows similar descriptions for two different action decisions.}
    \label{fig:teaser}
    \vspace{-3mm}
\end{figure}

Early approaches for zero-shot LLM-based VLN agents interpret the visual environment by utilizing offline Vision-Language Models (VLMs)~\cite{li2023blip, liu2023llava, wang2022internvideo} to convert visual images into the corresponding textual descriptions~\cite{zhou2024navgpt, long2024discuss, qiao2023march}. However, as shown in Fig.~\ref{fig:teaser}, these textual descriptions often provide very similar information when candidate images contain overlapping views, even if they are captured from different angles. 
More recently, MapGPT~\cite{chen2024mapgpt} processes multiple images simultaneously, directly feeding them into LLMs as input. This approach reduces redundancy in textual descriptions by leveraging visual differences, but this effort also remains limited when handling highly similar images, such as when both images depict \textit{``a kitchen''} in Fig.~\ref{fig:teaser}. 
Motivated by these challenges, we hypothesize that incorporating additional analogical reasoning processes is necessary to help the agent distinguish key features within the visually similar images while leveraging spatial information to discern their positional differences~(\textit{e.g.}, \textit{``slightly left''}).


To address the aforementioned challenge, we introduce a novel \textbf{analogical reasoning module} aimed at enhancing the LLM-based VLN agent’s \textbf{contextual understanding}, with a focus on both \textit{scene understanding from images} and \textit{spatial reasoning within the environment}. Our approach generates textual descriptions of the visual observations and leverages the power of language to describe differences between images, enabling the agent to capturing higher-level scene understanding and spatial relationships. 
Specifically, instead of treating candidate images as isolated inputs and prompting LLMs to generate independent visual descriptions, we leverage VLMs to compare multiple images and generate contextualized scene descriptions that highlight each image's distinctive features.
Furthermore, to strengthen the agent's spatial reasoning, we encourage it to systematically organize and interpret the spatial relationships between images. Unlike previous methods~\cite{zhou2023navgpt, pan2024langnav} that rely on rigid thresholds to define nuanced spatial concepts (\textit{e.g.}, strictly labeling directions as \textit{``slightly left"} or \textit{``further left"}), our approach enables the agent to distinguish these subtle differences through explicit visual comparisons and language-based reasoning. Specifically, we generate a detailed descriptive paragraph that explicitly captures the spatial relationships among the images based on raw spatial attributes such as rotation angles and distances.

We evaluate our method on the VLN mainstream benchmark Room-to-Room (R2R)~\cite{anderson2018vision} and REVERIE~\cite{qi2020reverie}. Experimental results demonstrate that incorporating our proposed analogical reasoning and spatial descriptions significantly improve navigation performance compared to using raw text or images alone. Furthermore, combining images with our proposed textual descriptions yields the best performance, highlighting the effectiveness of our descriptions in enhancing the agent's analogical reasoning, beyond what reasoning over visual input alone can provide.

%% file: sections/related_work.tex
\section{Related Works}
\myparagraph{Vision-and-Language Navigation (VLN)}  is a challenging embodied AI task that requires an agent to navigate in a photo-realistic environment by 
instructions~\cite{anderson2018vision, anderson2020rxr, qi2020reverie}. 
With the rise of foundation models, most VLN agents focus on integrating pre-trained models and generating large-scale datasets to enhance multi-modal representations~\cite{hamt, hop, tan2019learning, li2022envedit, wang2023scaling, wangbootstrapping, guhur2021airbert,  li2024panogen, zhang2023vln, zhang2022lovis, zhang2022explicit, zhang2021towards, zhang2024navhint}. 
Recently, LLMs and VLMs offer VLN a promising solution to mitigate domain-specific training constraints, particularly for zero-shot VLN agents~\cite{zhou2024navgpt, zhou2024navgpt2, chen2024mapgpt, long2023discuss, zhang2024navid, zheng2024towards, qiao2024llm, ma2025breaking}. However, current LLM-based VLN agents struggle with distinguishing visually similar scenes and exhibit limited spatial understanding. Our goal is to improve agents by addressing both challenges.



\noindent\textbf{Analogical Reasoning} is a cognitive process that involves comparing different entities to identify underlying structural similarities
~\cite{lovett2009solving, lovett2017modeling, huang2021diagnostic, grice1975logic, mitkov2022oxford, fried2022pragmatics}.
Recent work ~\cite{webb2023emergent, yuthought} leverages analogical reasoning, particularly within large language models (LLMs), to improve their understanding and reasoning capabilities in tasks requiring structural alignment or relational comparisons.
Analogical reasoning also facilitates comparisons between visual representations of objects observed from different camera views, leading to an improved global understanding of scenes~\cite{mitra2023one}.
Our work extends analogical reasoning to VLN tasks, enabling agents to compare discrete images, discern similarities and differences, and develop a global understanding of the environment.

%% file: sections/method.tex
\section{Methods}
\begin{figure*}
    \centering
\includegraphics[width=0.96\linewidth]{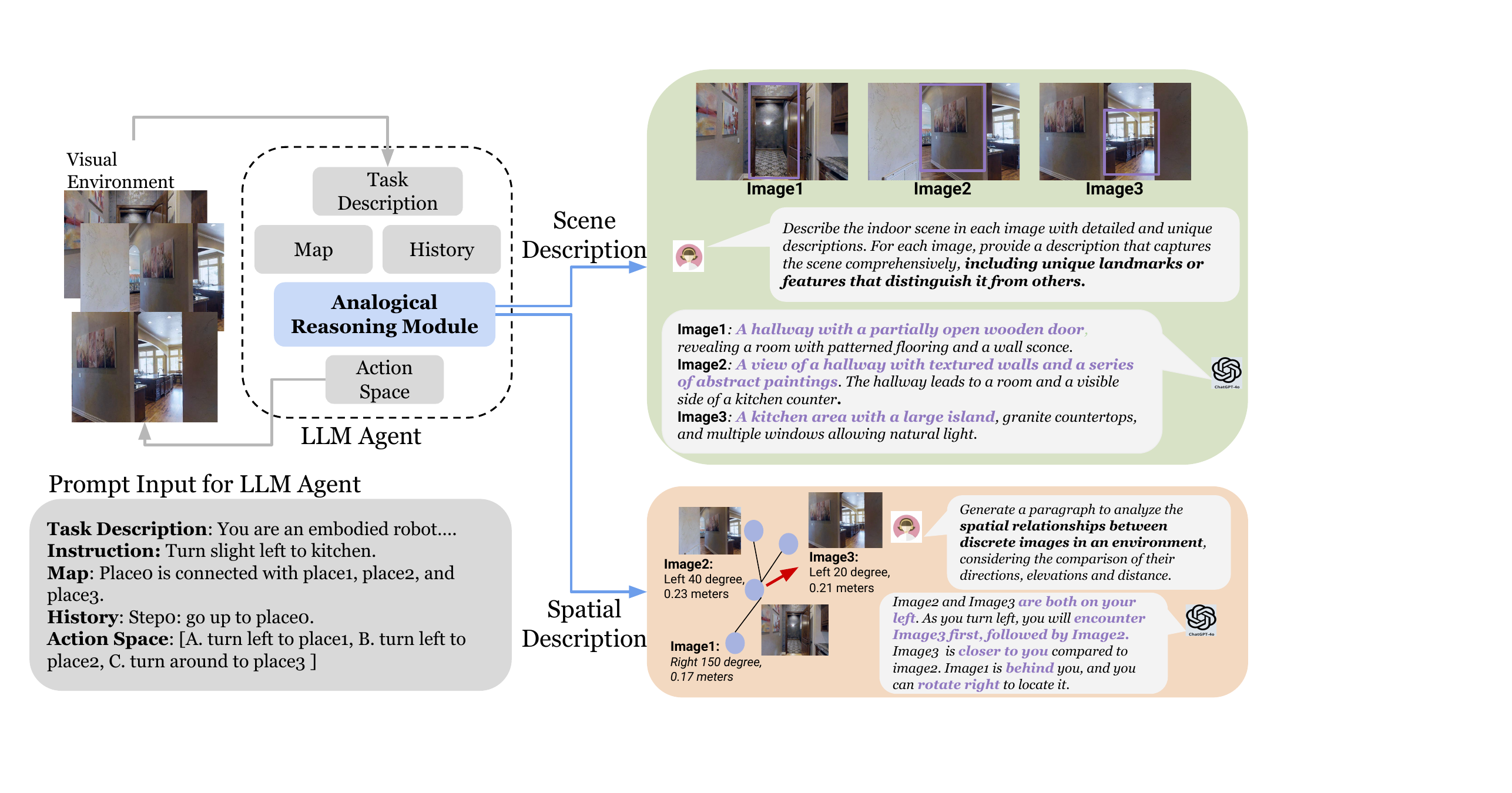}
    \caption{Model Architecture. Our method builds upon an LLM-based navigation agent that takes various prompt inputs~(\textcolor{lightgray}{\rule{0.4cm}{0.25cm}}). Built upon it, we introduce an analogical reasoning module composed of two components: analogical scene descriptions~(\textcolor{PaleSageGreen}{\rule{0.4cm}{0.25cm}}) and spatial descriptions (\textcolor{MutedApricot}{\rule{0.4cm}{0.25cm}}).}  
    \label{fig:main architecture}
\end{figure*}

In this section, we present our method, which builds upon MapGPT and aims to enhance analogical reasoning by using visual and spatial information of the environment.
The model architecture has been shown in Fig~\ref{fig:main architecture}.

\subsection{Task Formulation} In the VLN task, an agent receives a natural language instruction, denoted as $I$. At each navigation step, the agent perceives visual observations consisting of $n$ discrete images and selects one of these images as its action. The objective is to generate a trajectory (a sequence of images) that follows the given instruction.
To achieve this, the LLM-based VLN agent takes multiple sources of information as input, including instruction $I$, history $H_t$, topological map $M_t$, observation $O_t$, and action space $A_t$. The agent's decision-making process at step $t$ is formulated as:
\begin{equation}
    a_t = LLM(I, H_t, M_t, O_t, A_t),
\vspace{-3mm}
\end{equation}
where $a_t \in A_t$. As shown in Fig.~\ref{fig:main architecture}, the history includes previous step actions, capturing the sequence of movements. The map shows the connectivity graph between places~(images).
The action space is defined as a combination of direction and image (place), where the direction is determined based on both heading and elevation, including: \textit{go forward}, \textit{turn left/right/around}, and \textit{go up/down}.


\subsection{Analogical Reasoning Module}
As shown in Fig.~\ref{fig:main architecture}, our analogical reasoning module consists of two components: scene descriptions, which highlight differences between viewpoint images, and spatial descriptions, which capture spatial relationships between viewpoint images and emphasize differences relative to the agent's perspective.

\subsubsection{Scene Descriptions}
For different LLM-based VLN agents, one of the primary differences lies in how observations $O$ are represented. For instance, NavGPT~\cite{zhou2024navgpt} and DiscussNav~\cite{long2024discuss} utilize VLMs~(\textit{e.g.} BLIP-2~\cite{li2023blip}) to convert visual images into corresponding textual descriptions. While this approach enables language-driven navigation, it has a critical limitation: these textual descriptions treat each discrete image independently, disregarding contextual information across frames. 
However, a robust VLN agent should not only generate textual descriptions but also ensure that these descriptions encode contextual and relational differences across observations. To achieve this, we propose prompting LLMs to generate detailed visual descriptions while explicitly emphasizing the distinguishing features between different observations, which is formally denoted as follows:
\begin{equation}
    \small
OT_1,OT_2,\dots,OT_n=\texttt{LLM}\big(\texttt{Prompt}(O_1,O_2,\dots,O_n)\big),
\end{equation}
where \texttt{Prompt} is instructions designed to guide the LLMs in generating an analogical analysis of the input. $OT_i$ represents the textual description of the corresponding image $O_i$.

We illustrate our approach with an example in Fig.~\ref{fig:main architecture}, where the prompts are demonstrated alongside the corresponding textual descriptions generated for the given images.
Our method strategically prompts LLMs to identify distinguishing landmarks that differentiate each image from the others. As a result, the opening sentence of each visual description explicitly highlights these unique features, ensuring a clear comparative distinction.
For instance, in Image 1, the description emphasizes a hallway featuring a wooden door, whereas in Image 2, the focus shifts to a hallway with paintings, leading to a room and a kitchen counter. Meanwhile, Image 3 directs attention to a kitchen area centered around a large island.
By emphasizing analogical attributes rather than describing each image in isolation, our approach enhances contextual understanding and strengthens the model’s ability to discern subtle yet critical differences between visually similar scenes.

\subsubsection{Spatial Descriptions}
A key challenge for the LLM-based VLN agent is effectively representing the spatial structure of its visual environment. In current methods~\cite{chen2024mapgpt, pan2024langnav}, the action space is highly discretized, allowing only a generic directional instruction, \textit{e.g.}, \textit{``turn left''}, which are defined based on fixed angle thresholds.
This classification overlooks subtle yet crucial spatial distinctions, such as differentiating a slight 5-degree turn from a more significant 30-degree rotation. Consequently, the agent cannot accurately interpret or respond to nuanced navigation instructions like \textit{``turn slightly left''}, 
causing imprecise action decisions.


A straightforward approach to addressing this limitation is to directly provide raw heading and elevation values. For example, rather than the ambiguous instruction \textit{``turn left''} the action space could specify \textit{``turn left $5$ degrees''}.  However, our experiments~(Appendix~\ref{different spatial}) reveal that the agent struggles to effectively comprehend and integrate this fine-grained spatial information, suggesting that merely providing numerical orientation values is insufficient for enhancing its spatial reasoning.

To address this challenge, we draw inspiration from the approach of obtaining analogical scene descriptions from images and extend it to spatial understanding. Our approach focuses on constructing a structured contextual representation that captures spatial relationships across discrete images. Fig.~\ref{fig:main architecture} illustrates our designed prompts for describing spatial relationships. We begin by computing the spatial relation, including the relative rotational angle (\textit{e.g.}, ``left by 20 degrees'') and the relative distance (\textit{e.g.}, ``0.21 meters''; note that MapGPT ignores distance). These computed attributes are then incorporated into a structured prompt that guides the LLMs to generate a detailed paragraph analyzing the spatial relationships. The generated description explicitly considers directional comparisons, elevation differences, and distance variations, ensuring a comprehensive understanding of the spatial context. 
We provide full prompts in the Appendix~\ref{full prompts}.
We denote the generated spatial description as $S$, and our enhanced LLM agent’s decision-making process is finally defined as follows:
\begin{equation}
    a_t = LLM(I, H_t, M_t, \{O_t,OT_t\}, S_t, A_t),
\vspace{-2mm}
\end{equation} 
where $\{O_t, OT_t\}$ indicates that our agent can flexibly take either the image, its corresponding scene description, or both as inputs.



%% file: sections/experiment.tex
\section{Experiments}

\begin{table}[]
    \centering
    \resizebox{\columnwidth}{!}{%
    \begin{tabular}{c c c c c c}
    \toprule
         & \textbf{Methods}  & \textbf{NE}$\downarrow$ & \textbf{OSR} $\uparrow$ & \textbf{SR}$\uparrow$ & \textbf{SPL}$\uparrow$  \\
        \hline
       \multirow{4}{*}{R2R} & NavGPT~(with GPT-4) &    
       $6.46$ & $42$ & $34$ & $29$ \\
        & MapGPT~(with GPT-4) & $6.29$ & $61.6$ & $38.8$ & $25.8$ \\
        & MapGPT~(with GPT-4V) & $5.63$ & $57.9$ & $43.7$ & $34.8$ \\
        & MapGPT~(with GPT-4o) & $5.31$ & $56.9$  & $43.8$ & $36.5$  \\
         & Ours~(with GPT-4o) & $\mathbf{4.79}$ & $\mathbf{65.7}$ &$\mathbf{49.5}$ & $\mathbf{42.5}$ \\
         \midrule
        \multirow{2}{*}{REVERIE} & MapGPT~(with GPT-4o) & - & $43.33$ &  $30.00$ & $22.58$\\
        & Ours~(with GPT-4o) & - & $\mathbf{50.00}$ &  $\mathbf{33.16}$& $\mathbf{26.09}$\\
    \bottomrule
    \end{tabular}
    }
    \caption{Results on the validation unseen set of the R2R dataset. We implement our method solely on GPT-4o~\cite{openai2024gpt4o}, as GPT-4V has been deprecated.}
    \label{tab:unseen}
\end{table}

\subsection{Experimental Settings}
\noindent\textbf{Datasets and Evaluation Metrics.} 
We evaluate our method on the R2R~\cite{anderson2018vision} and REVERIE~\cite{qi2020reverie}, two standard benchmarks for VLN. 
We also follow MapGPT in conducting evaluations on a sampled subset of the R2R dataset for our ablation study, including $72$ scenarios and $216$ examples. Please also check our released code~\footnote{https://github.com/zhangyuejoslin/VLN-Analogical-Reasoning}.

Three main metrics are used to evaluate navigation performance: (1) Navigation Error (NE): the mean of the shortest path distance between the agent's final position and the goal destination. (2) Success Rate (SR): the percentage of the predicted final position being within 3 meters from the goal destination. (3) Success Rate Weighted Path Length (SPL): normalizes success rate by trajectory length. (4) Oracle Success Rate: the agent passes by or stops at a graph node
within 3 meters to the destination.

\begin{table}[t]
    \centering
    \resizebox{\columnwidth}{!}{%
    \begin{tabular}{c c c c c c c}
    \toprule
    \textbf{Methods} & \# & \textbf{Image} & \textbf{Text} & \textbf{GPT} & \textbf{SR}$\uparrow$ & \textbf{SPL}$\uparrow$ \\
    \hline
    NavGPT & $1$ & - & BLIP-2 & GPT-3.5 & $16.7$ & $13.0$ \\
    \hline
    \multirow{6}{*}{MapGPT} 
        & $2$ & - & BLIP-2 & GPT-4 & $41.2$ & $25.4$ \\
        & $3$ & - & BLIP-2 & GPT-4o &  $38.5$ & $26.9$ \\
        & $4$ & - &GPT-4o & GPT-4o & $45.6$ & $36.2$ \\
        & $5$ & \ding{52} & - & Qwen2.5-VL-7B & $30.0$ & $26.6$ \\
        & $6$ & \ding{52} & - &  GPT-4v & $47.7$ & $38.1$\\
        & $7$ & \ding{52} & - &  GPT-4o-05-13 & $41.2$ & $35.1$ \\
        & $8$ & \ding{52} & - &  GPT-4o & $47.7$ & $38.7$\\
    \hline
    \multirow{3}{*}{Ours} 
        & $9$ & - & GPT-4o (SI) & GPT-4o & $48.2$ & $36.2$ \\
        & $10$ & - & GPT-4o (SP) & GPT-4o & $47.4$ & $36.2$ \\
        & $11$ & - & GPT-4o (SI+SP) & GPT-4o & $\mathbf{50.0}$ & $36.4$ \\
        & $12$ &\ding{52} & Qwen2.5-VL-7B (SI+SP) & Qwen2.5-VL-7B & $32.4$ & $28.1$ \\
        & $13$ & \ding{52} & GPT-4o (SI+SP) & GPT-4o & $\mathbf{50.0}$ & $\mathbf{40.2}$ \\
    \bottomrule
    \end{tabular}
    }
    \caption{Ablation Results on 72 diverse scenes~\cite{chen2024mapgpt} from the R2R dataset. All GPT-4o versions are from the 08-06 release, except GPT-4o-05-13. SI: scene descriptions; SP: spatial descriptions.}
    \label{tab:results_r2r}
\end{table}

\noindent\textbf{Implementation Details.} 
We utilize GPT-4o-08-06 as the backbone for our LLM-based agent, given that GPT-4V has been deprecated. In this work, we employ GPT-4o-08-06 as the backbone for our LLM-based agent, as GPT-4V has been deprecated. MapGPT reports its results using GPT-4o-05-03, but our implementation with GPT-4o-08-06 achieves better performance~(around $6\%$ on success rate).
To ensure deterministic outputs, we set the temperature to $0$. Additionally, we constrain the agent's decision-making process by limiting the maximum number of generated actions to $15$ and the maximum token output from GPT to $2000$.

\subsection{Experimental Results} 
Table~\ref{tab:unseen} shows the results on the R2R and REVERIE unseen dataset, demonstrating that our method significantly enhances the baselines, achieving around $4-6\%$ improvement in both SR and SPL.
Table~\ref{tab:results_r2r} demonstrates our results on 72 diverse scenes. We compare our approach against other LLM-based agents, varying the image input, text input, and GPT backbones.
We highlight several key findings:
\textbf{First}, both scene and spatial descriptions contribute incrementally to navigation performance. Using row \#$4$ as a baseline, we observe performance improvements when independently adding scene descriptions (row \#$9$) or spatial descriptions (row \#$10$), with further gains achieved when combining both types of descriptions (row \#$11$).
\textbf{Second}, our analogical reasoning descriptions also enhance the agent’s ability to reason over visual inputs. Comparing row \#$8$ and row \#$13$ shows that while raw images contain all necessary visual information, our structured text-based descriptions provide complementary high-level reasoning, leading to improved performance.
\textbf{Third}, our approach generalizes across different backbones. For example, applying our method to another backbone model of Qwen2.5 shows consistent gains~(comparing Row \#$5$ with Row \#$12$) confirms that our approach enhances performance beyond GPT-4o.
\textbf{Fourth}, stronger captioners help better scene understanding. For instance, BLIP-2 (row \#$3$) underperforms significantly compared to GPT-4o (row \#$4$), showing the importance of a more advanced captioning model for generating informative text inputs.
\textbf{Fifth}, including images significantly improves SPL. As shown in row \#$11$ and row \#$13$, visual inputs help the agent ground non-salient target objects directly, reducing the need for extra exploration and shortening paths (higher SPL). Without images, the agent requires additional steps to locate less visually prominent targets through textual descriptions alone, resulting in longer paths but ultimately similar SR.

\subsection{Qualitative Examples}
\label{qualitative example}
In addition to the quantitative results, Fig.~\ref{fig:scene qualitative} and Fig.~\ref{fig:spatial qualitative} in Appendix~\ref{qualitative example} 
present two qualitative examples illustrating the effectiveness of the proposed analogical scene and spatial descriptions. In Fig.~\ref{fig:scene qualitative}, the scene descriptions generated by BLIP-2 and GPT-4o are highly similar despite the visual differences between the scenes. Even for GPT-4o, across three images, the descriptions primarily focus on the general scene, referring to an \textit{``ornate chapel interior''} without providing distinguishing details. In contrast, our method emphasizes different aspects of each image: for example, Image 1 highlights \textit{``the confessional booth''}, Image 2 focuses on \textit{``the benches''}, and Image 3 emphasizes \textit{``the grand altar''}. These distinct descriptions enable the agent to accurately select Image 2, which aligns with the given instruction.
Furthermore, in Fig~\ref{fig:spatial qualitative}, We present an example demonstrating the effectiveness of spatial descriptions. In this case, both Image 4 and Image 5 contain an entranceway. However, our approach encourages the agent to infer that less left/right rotation corresponds to a direction closer to forward. As a result, the agent correctly reasons that Image 5 is better aligned with the instruction \textit{``walk to''}.

%% file: sections/conclusion.tex
\section{Conclusion}
In this paper, we propose enhancing the contextual understanding of LLM-based VLN agents by generating analogical scene and spatial descriptions. We encourage the agent to compare images from different perspectives and help the agent construct a structured spatial understanding of the environment. We evaluate our method on the R2R dataset and demonstrate that our approach significantly improves performance compared to LLM-based navigation agents. 

\section{Acknowledgment}
We thank the reviewers and area chairs for their
helpful feedback. This project is partially supported by the Office of Naval Research (ONR) grant N00014-23-1-2417. Any opinions, findings, and conclusions or recommendations expressed in this material are those of the authors and do not necessarily reflect the views of Office of Naval Research.

\section{Limitation}
Despite the significant improvement in navigation performance achieved by our analogical reasoning descriptions, several limitations remain. First, the quality of the generated descriptions heavily depends on the underlying language model, which may introduce biases or hallucinations that could impact decision-making. Second, the process of generating analogical descriptions adds an additional computational step, potentially increasing processing costs compared to direct image-based navigation.

%% file: sections/appendix.tex
\section{Appendix}
\label{sec:appendix}


\subsection{Prompts for Spatial Descriptions}
\label{full prompts}
Fig.~\ref{fig:spatial description} shows the prompts we used for spatial descriptions.

\subsection{Different Strategies for Spatial Reasoning}
\label{different spatial}
We conduct experiments to examine how different spatial reasoning strategies impact navigation performance. Intuitively, enabling an agent to understand nuanced spatial concepts can be achieved by explicitly incorporating varying degrees of rotation into its action space. For example, the agent's action space is more precisely defined, such as \textit{``turn 5 degrees left''}. However, our results reveal that introducing fine-grained rotational actions leads to a slight decline in navigation performance~(row \#$2$ in Table.~\ref{tab:experiment for different}). This suggests that VLN agents struggle to effectively structure spatial information when relying solely on numerical rotations degrees.
To address this, we propose generating descriptive paragraphs that systematically capture spatial relationships between images. Empirical results demonstrate that our approach enhances navigation performance compared to directly using numerical values into the action space~(\#$3$ in Table.~\ref{tab:experiment for different}).

\begin{figure}[t]
\centering
\small
\begin{tcolorbox}[title=Prompt for Spatial Descriptions, colback=gray!5, colframe=gray!80, fonttitle=\bfseries]

Generate a paragraph to analyze the spatial relationships between discrete images in an environment, considering the comparision of their directions, elevations and distance. The input consists of images with specific angles and distances relative to a central point. Here are some rules to follow:
Angles between 120 to 240 degree to the left or right indicate behind or around.
Angles equals 180 degrees indicate direct behind.
Less angles rotation degrees to the left or right indicate closer to the forward direction. For example, Given places along with their spatial information: Place0 is to my right 180.0 degrees and up 30.0 degrees, positioned 0.21 meters away, Place 2 is to my right 60.0 degrees and up 30.0 degrees, positioned 0.21 meters away Place 3 is to my right 90.0 degrees and up 30.0 degrees, positioned 0.18 meters away. Place 4 is to my right 90.0 degrees, positioned 0.05 meters away. Please generate a descriptive paragraph explaining the spatial relationships and navigation steps to these images. For example:
        ``To navigate to Image0, Image2, and Image3, you need to move upward. As you turn right, you will encounter Image2 first, followed by Image3, and finally Image0, which is directly behind you. Image4 is in the same direction as Image3, but Image3 requires looking up while Image4 does not. Additionally, Image4 is very close to you.''
        Output the response in JSON format with the key 'environmental analysis.' 
        "
\end{tcolorbox}
\caption{Prompt for Spatial Descriptions.}
\label{fig:spatial description}
\end{figure}

\begin{table}[h]
    \centering
    \resizebox{0.8\columnwidth}{!}{%
    \begin{tabular}{c c c  c c}
    \toprule
    \# &    \textbf{Methods}  & \textbf{SR}$\uparrow$ & \textbf{SPL}$\uparrow$  \\
        \hline
      $1$ & MapGPT & $47.7$ & $38.7$ \\
      $2$ & +spatial attributes &  $46.8$ & $37.9$ \\
        \hline
       $3$ &  +spatial descriptions  & $\mathbf{49.1}$ & $\mathbf{39.3}$ \\
    \bottomrule
    \end{tabular}
    }
    \caption{Different strategies for spatial reasoning.}
    \vspace{-3mm}
    \label{tab:experiment for different}
\end{table}

\begin{figure*}[h]
    \centering
    \includegraphics[width=\linewidth]{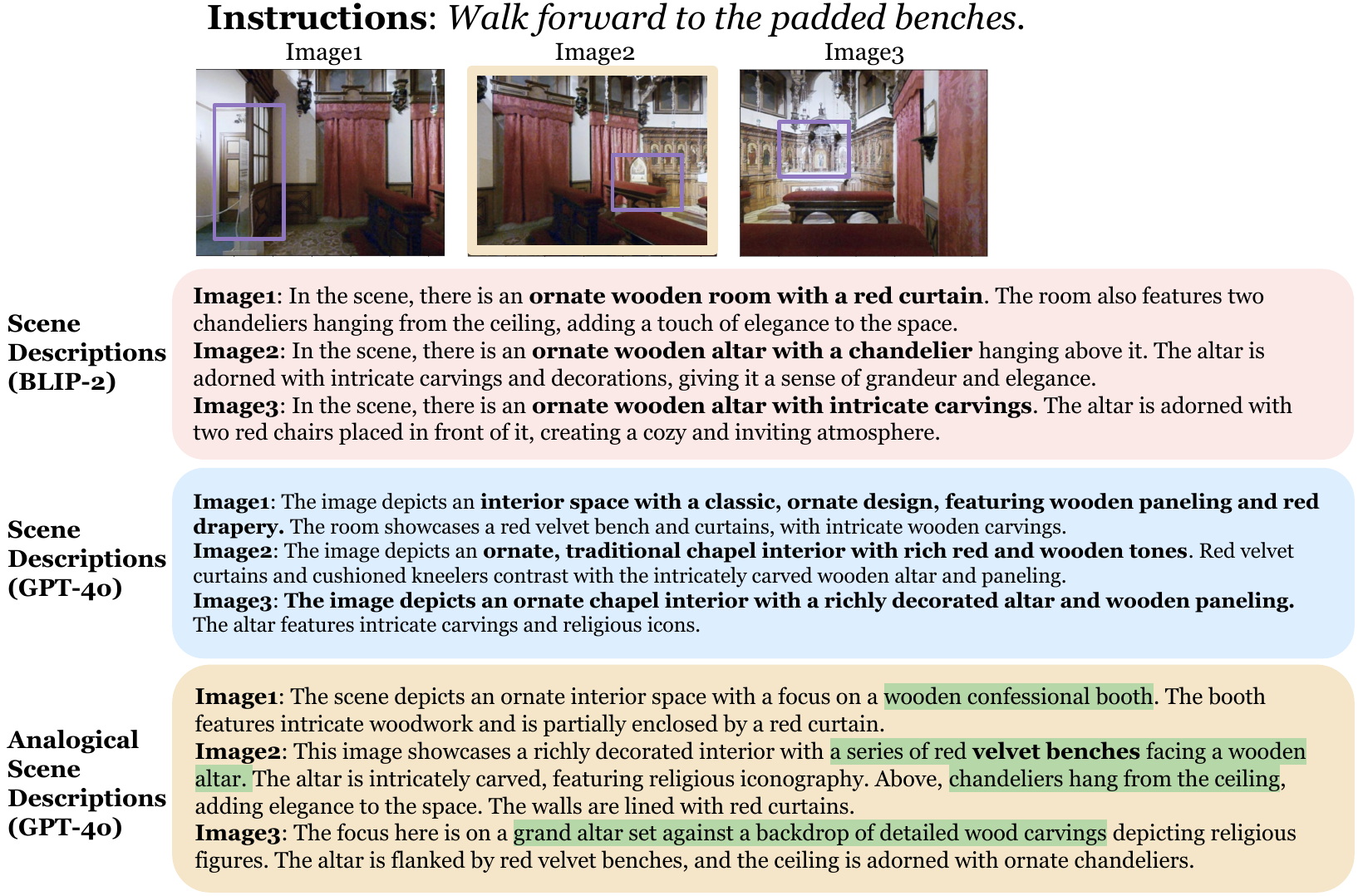}
    \caption{A qualitative example demonstrating the effectiveness of analogical scene descriptions. We collect scene descriptions from both BLIP-2 and GPT-4o. Bold text highlights similar descriptions across images, while text in green boxes represents our generated analogical scene descriptions, each emphasizing different aspects.}
    \label{fig:scene qualitative}
\end{figure*}

\begin{figure*}[t]
    \centering
    \includegraphics[width=0.9\linewidth]{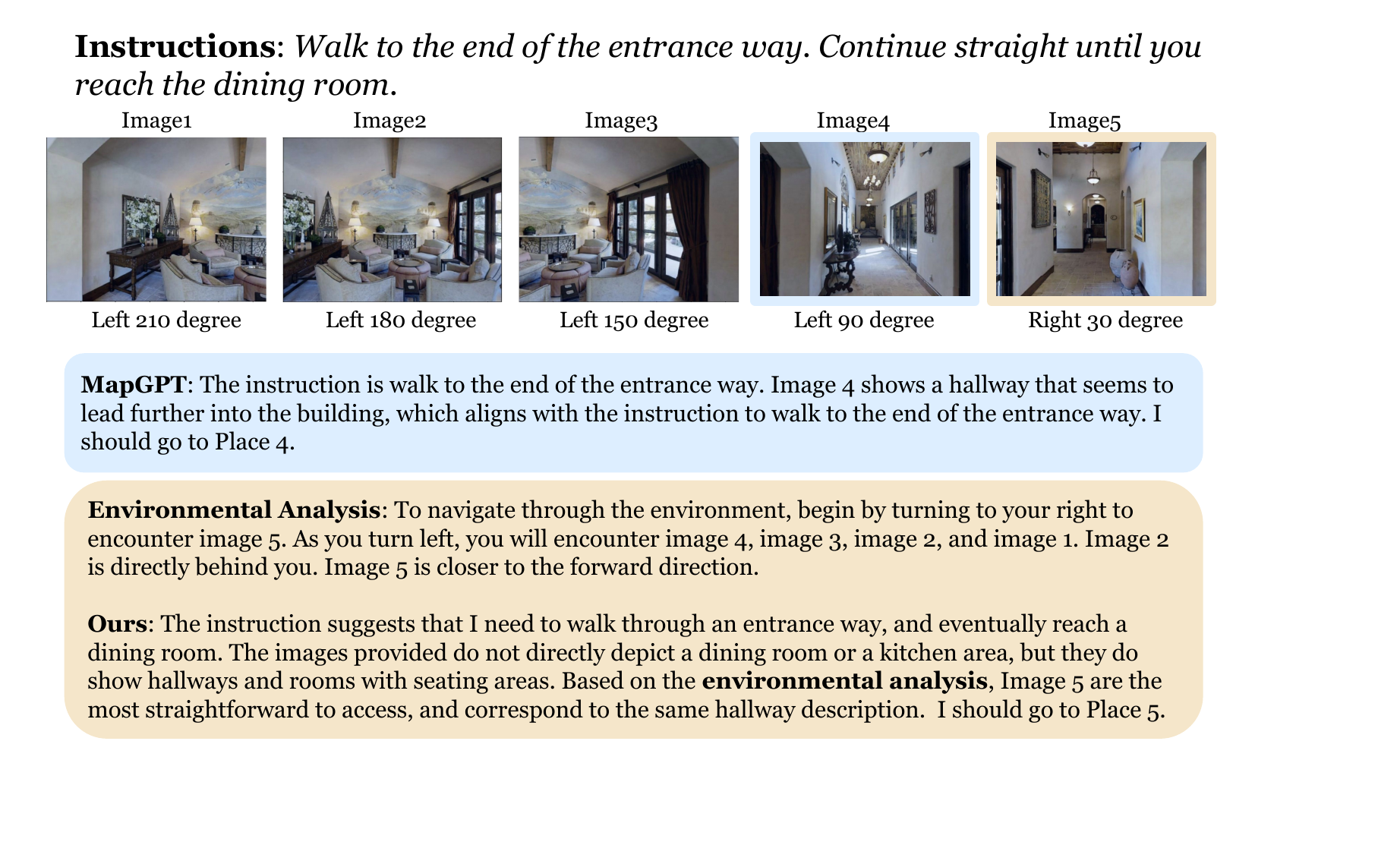}
    \caption{A qualitative example illustrating the effectiveness of our spatial descriptions. The agent successfully identified Place5 based on its relative position, as it is closer to the forward direction than other images and better aligned with the instruction \textit{``walk to''} compared to Place4, which requires a significant left turn.}
    \label{fig:spatial qualitative}
\end{figure*}

\subsection{Failure Cases Discussion}
One type of failure come from hallucination issues, these are inherently related to the specific LLM used; for instance, we observed more frequent hallucinations in BLIP compared to GPT-4o.
Another important type of failure case we noticed occurs when the agent overly emphasizes intermediate landmarks. For example, given the instruction ``walk past the dining table and head all the way up the stairs," the optimal action would be to directly select an image clearly showing the stairs, treating the dining table merely as a passing object. However, the model occasionally prioritizes the intermediate landmark (``the dining table''), potentially directing its attention away from the main goal. To address this, we recognize that analogical reasoning alone may not fully resolve this issue, and suggest that incorporating a hierarchical reasoning process to explicitly prioritize target landmarks could improve navigation performance.

%% file: acl_latex.bbl
\begin{thebibliography}{46}
\providecommand{\natexlab}[1]{#1}

\bibitem[{Anderson et~al.(2018)Anderson, Wu, Teney, Bruce, Johnson, S{\"u}nderhauf, Reid, Gould, and Van Den~Hengel}]{anderson2018vision}
Peter Anderson, Qi~Wu, Damien Teney, Jake Bruce, Mark Johnson, Niko S{\"u}nderhauf, Ian Reid, Stephen Gould, and Anton Van Den~Hengel. 2018.
\newblock Vision-and-language navigation: Interpreting visually-grounded navigation instructions in real environments.
\newblock In \emph{Proceedings of the IEEE conference on computer vision and pattern recognition}, pages 3674--3683.

\bibitem[{Chen et~al.(2024)Chen, Lin, Xu, Chai, Liang, and Wong}]{chen2024mapgpt}
Jiaqi Chen, Bingqian Lin, Ran Xu, Zhenhua Chai, Xiaodan Liang, and Kwan-Yee Wong. 2024.
\newblock Mapgpt: Map-guided prompting with adaptive path planning for vision-and-language navigation.
\newblock In \emph{Proceedings of the 62nd Annual Meeting of the Association for Computational Linguistics (Volume 1: Long Papers)}, pages 9796--9810.

\bibitem[{Chen et~al.(2021)Chen, Guhur, Schmid, and Laptev}]{hamt}
Shizhe Chen, Pierre{-}Louis Guhur, Cordelia Schmid, and Ivan Laptev. 2021.
\newblock History aware multimodal transformer for vision-and-language navigation.
\newblock In \emph{Advances in Neural Information Processing Systems}, pages 5834--5847.

\bibitem[{Fried et~al.(2022)Fried, Tomlin, Hu, Patel, and Nematzadeh}]{fried2022pragmatics}
Daniel Fried, Nicholas Tomlin, Jennifer Hu, Roma Patel, and Aida Nematzadeh. 2022.
\newblock Pragmatics in language grounding: Phenomena, tasks, and modeling approaches.
\newblock \emph{arXiv preprint arXiv:2211.08371}.

\bibitem[{Grice(1975)}]{grice1975logic}
Herbert~P Grice. 1975.
\newblock Logic and conversation.
\newblock In \emph{Speech acts}, pages 41--58. Brill.

\bibitem[{Guhur et~al.(2021)Guhur, Tapaswi, Chen, Laptev, and Schmid}]{guhur2021airbert}
Pierre-Louis Guhur, Makarand Tapaswi, Shizhe Chen, Ivan Laptev, and Cordelia Schmid. 2021.
\newblock Airbert: In-domain pretraining for vision-and-language navigation.
\newblock In \emph{Proceedings of the IEEE/CVF International Conference on Computer Vision}, pages 1634--1643.

\bibitem[{Guo et~al.(2025)Guo, Song, Zhang, Liu, and Liu}]{guo2025rethinking}
Xiao Guo, Xiufeng Song, Yue Zhang, Xiaohong Liu, and Xiaoming Liu. 2025.
\newblock Rethinking vision-language model in face forensics: Multi-modal interpretable forged face detector.
\newblock In \emph{Proceedings of the Computer Vision and Pattern Recognition Conference}, pages 105--116.

\bibitem[{Huang et~al.(2021)Huang, Zhu, Sun, Choi, Tan, and Lim}]{huang2021diagnostic}
Ziqi Huang, Hongyuan Zhu, Ying Sun, Dongkyu Choi, Cheston Tan, and Joo-Hwee Lim. 2021.
\newblock A diagnostic study of visual question answering with analogical reasoning.
\newblock In \emph{2021 IEEE International Conference on Image Processing (ICIP)}, pages 2463--2467. IEEE.

\bibitem[{Ku et~al.(2020)Ku, Anderson, Patel, Ie, and Baldridge}]{anderson2020rxr}
Alexander Ku, Peter Anderson, Roma Patel, Eugene Ie, and Jason Baldridge. 2020.
\newblock Room-across-room: Multilingual vision-and-language navigation with dense spatiotemporal grounding.
\newblock In \emph{Proceedings of the 2020 Conference on Empirical Methods in Natural Language Processing (EMNLP)}, pages 4392--4412.

\bibitem[{Li and Bansal(2024)}]{li2024panogen}
Jialu Li and Mohit Bansal. 2024.
\newblock Panogen: Text-conditioned panoramic environment generation for vision-and-language navigation.
\newblock In \emph{Conference on Neural Information Processing Systems}, volume~36.

\bibitem[{Li et~al.(2022)Li, Tan, and Bansal}]{li2022envedit}
Jialu Li, Hao Tan, and Mohit Bansal. 2022.
\newblock Envedit: Environment editing for vision-and-language navigation.
\newblock In \emph{Proceedings of the IEEE/CVF Conference on Computer Vision and Pattern Recognition}, pages 15407--15417.

\bibitem[{Li et~al.(2023)Li, Li, Savarese, and Hoi}]{li2023blip}
Junnan Li, Dongxu Li, Silvio Savarese, and Steven Hoi. 2023.
\newblock Blip-2: Bootstrapping language-image pre-training with frozen image encoders and large language models.
\newblock In \emph{International conference on machine learning}, pages 19730--19742. PMLR.

\bibitem[{Liu et~al.(2023)Liu, Li, Wu, and Lee}]{liu2023llava}
Haotian Liu, Chunyuan Li, Qingyang Wu, and Yong~Jae Lee. 2023.
\newblock Visual instruction tuning.
\newblock In \emph{NeurIPS}.

\bibitem[{Long et~al.(2024{\natexlab{a}})Long, Li, Cai, and Dong}]{long2024discuss}
Yuxing Long, Xiaoqi Li, Wenzhe Cai, and Hao Dong. 2024{\natexlab{a}}.
\newblock Discuss before moving: Visual language navigation via multi-expert discussions.
\newblock In \emph{2024 IEEE International Conference on Robotics and Automation (ICRA)}, pages 17380--17387. IEEE.

\bibitem[{Long et~al.(2024{\natexlab{b}})Long, Li, Cai, and Dong}]{long2023discuss}
Yuxing Long, Xiaoqi Li, Wenzhe Cai, and Hao Dong. 2024{\natexlab{b}}.
\newblock Discuss before moving: Visual language navigation via multi-expert discussions.
\newblock In \emph{2024 IEEE International Conference on Robotics and Automation (ICRA)}, pages 17380--17387. IEEE.

\bibitem[{Lovett and Forbus(2017)}]{lovett2017modeling}
Andrew Lovett and Kenneth Forbus. 2017.
\newblock Modeling visual problem solving as analogical reasoning.
\newblock \emph{Psychological review}, 124(1):60.

\bibitem[{Lovett et~al.(2009)Lovett, Tomai, Forbus, and Usher}]{lovett2009solving}
Andrew Lovett, Emmett Tomai, Kenneth Forbus, and Jeffrey Usher. 2009.
\newblock Solving geometric analogy problems through two-stage analogical mapping.
\newblock \emph{Cognitive science}, 33(7):1192--1231.

\bibitem[{Ma et~al.(2025)Ma, Zhang, Wang, and Kordjamshidi}]{ma2025breaking}
Tianyi Ma, Yue Zhang, Zehao Wang, and Parisa Kordjamshidi. 2025.
\newblock Breaking down and building up: Mixture of skill-based vision-and-language navigation agents.
\newblock \emph{arXiv preprint arXiv:2508.07642}.

\bibitem[{Mitkov(2022)}]{mitkov2022oxford}
Ruslan Mitkov. 2022.
\newblock \emph{The Oxford handbook of computational linguistics}.
\newblock Oxford university press.

\bibitem[{Mitra et~al.(2023)Mitra, Anwar, Corona, Klein, Darrell, and Thomason}]{mitra2023one}
Chancharik Mitra, Abrar Anwar, Rodolfo Corona, Dan Klein, Trevor Darrell, and Jesse Thomason. 2023.
\newblock Which one? leveraging context between objects and multiple views for language grounding.
\newblock \emph{arXiv preprint arXiv:2311.06694}.

\bibitem[{OpenAI(2024)}]{openai2024gpt4o}
OpenAI. 2024.
\newblock \href {https://openai.com/index/hello-gpt-4o/} {Hello gpt-4o}.

\bibitem[{Pan et~al.(2024)Pan, Panda, Jin, Feris, Oliva, Isola, and Kim}]{pan2024langnav}
Bowen Pan, Rameswar Panda, SouYoung Jin, Rogerio Feris, Aude Oliva, Phillip Isola, and Yoon Kim. 2024.
\newblock Langnav: Language as a perceptual representation for navigation.
\newblock In \emph{Findings of the Association for Computational Linguistics: NAACL 2024}, pages 950--974.

\bibitem[{Qi et~al.(2020)Qi, Wu, Anderson, Wang, Wang, Shen, and Hengel}]{qi2020reverie}
Yuankai Qi, Qi~Wu, Peter Anderson, Xin Wang, William~Yang Wang, Chunhua Shen, and Anton van~den Hengel. 2020.
\newblock Reverie: Remote embodied visual referring expression in real indoor environments.
\newblock In \emph{Proceedings of the IEEE/CVF Conference on Computer Vision and Pattern Recognition}, pages 9982--9991.

\bibitem[{Qiao et~al.(2024)Qiao, Liu, Liu, Liu, and Wu}]{qiao2024llm}
Yanyuan Qiao, Qianyi Liu, Jiajun Liu, Jing Liu, and Qi~Wu. 2024.
\newblock Llm as copilot for coarse-grained vision-and-language navigation.
\newblock In \emph{European Conference on Computer Vision}, pages 459--476. Springer.

\bibitem[{Qiao et~al.(2022)Qiao, Qi, Hong, Yu, Wang, and Wu}]{hop}
Yanyuan Qiao, Yuankai Qi, Yicong Hong, Zheng Yu, Peng Wang, and Qi~Wu. 2022.
\newblock {HOP:} history-and-order aware pre-training for vision-and-language navigation.
\newblock In \emph{Proceedings of the IEEE/CVF conference on Computer Vision and Pattern Recognition}, pages 8524--8537.

\bibitem[{Qiao et~al.(2023)Qiao, Qi, Yu, Liu, and Wu}]{qiao2023march}
Yanyuan Qiao, Yuankai Qi, Zheng Yu, Jing Liu, and Qi~Wu. 2023.
\newblock March in chat: Interactive prompting for remote embodied referring expression.
\newblock In \emph{Proceedings of the IEEE/CVF International Conference on Computer Vision}, pages 15758--15767.

\bibitem[{Tan et~al.(2019)Tan, Yu, and Bansal}]{tan2019learning}
Hao Tan, Licheng Yu, and Mohit Bansal. 2019.
\newblock Learning to navigate unseen environments: Back translation with environmental dropout.
\newblock In \emph{Proceedings of the 2019 Conference of the North American Chapter of the Association for Computational Linguistics: Human Language Technologies, Volume 1 (Long and Short Papers)}, pages 2610--2621.

\bibitem[{Wang et~al.(2022)Wang, Li, Li, He, Huang, Zhao, Zhang, Xu, Liu, Wang et~al.}]{wang2022internvideo}
Yi~Wang, Kunchang Li, Yizhuo Li, Yinan He, Bingkun Huang, Zhiyu Zhao, Hongjie Zhang, Jilan Xu, Yi~Liu, Zun Wang, et~al. 2022.
\newblock Internvideo: General video foundation models via generative and discriminative learning.
\newblock \emph{arXiv preprint arXiv:2212.03191}.

\bibitem[{Wang et~al.()Wang, Li, Hong, Li, Li, Yu, Wang, Qiao, Wang, Bansal et~al.}]{wangbootstrapping}
Zun Wang, Jialu Li, Yicong Hong, Songze Li, Kunchang Li, Shoubin Yu, Yi~Wang, Yu~Qiao, Yali Wang, Mohit Bansal, et~al.
\newblock Bootstrapping language-guided navigation learning with self-refining data flywheel.
\newblock In \emph{The Thirteenth International Conference on Learning Representations}.

\bibitem[{Wang et~al.(2023)Wang, Li, Hong, Wang, Wu, Bansal, Gould, Tan, and Qiao}]{wang2023scaling}
Zun Wang, Jialu Li, Yicong Hong, Yi~Wang, Qi~Wu, Mohit Bansal, Stephen Gould, Hao Tan, and Yu~Qiao. 2023.
\newblock Scaling data generation in vision-and-language navigation.
\newblock In \emph{Proceedings of the IEEE/CVF International Conference on Computer Vision}, pages 12009--12020.

\bibitem[{Webb et~al.(2023)Webb, Holyoak, and Lu}]{webb2023emergent}
Taylor Webb, Keith~J Holyoak, and Hongjing Lu. 2023.
\newblock Emergent analogical reasoning in large language models.
\newblock \emph{Nature Human Behaviour}, 7(9):1526--1541.

\bibitem[{Yu et~al.()Yu, He, and Ying}]{yuthought}
Junchi Yu, Ran He, and Zhitao Ying.
\newblock Thought propagation: An analogical approach to complex reasoning with large language models.
\newblock In \emph{The Twelfth International Conference on Learning Representations}.

\bibitem[{Yu et~al.(2025)Yu, Yoon, and Bansal}]{yu2024crema}
Shoubin Yu, Jaehong Yoon, and Mohit Bansal. 2025.
\newblock Crema: Generalizable and efficient video-language reasoning via multimodal modular fusion.
\newblock \emph{ICLR}.

\bibitem[{Zhang et~al.(2024{\natexlab{a}})Zhang, Wang, Xu, Zhou, Hong, Fang, Wu, Zhang, and He}]{zhang2024navid}
Jiazhao Zhang, Kunyu Wang, Rongtao Xu, Gengze Zhou, Yicong Hong, Xiaomeng Fang, Qi~Wu, Zhizheng Zhang, and Wang He. 2024{\natexlab{a}}.
\newblock Navid: Video-based vlm plans the next step for vision-and-language navigation.
\newblock In \emph{Robotics: Science and Systems (RSS)}.

\bibitem[{Zhang et~al.(2024{\natexlab{b}})Zhang, Colman, Guo, Shahriyari, and Bharaj}]{zhang2024common}
Yue Zhang, Ben Colman, Xiao Guo, Ali Shahriyari, and Gaurav Bharaj. 2024{\natexlab{b}}.
\newblock Common sense reasoning for deepfake detection.
\newblock In \emph{European Conference on Computer Vision}, pages 399--415. Springer.

\bibitem[{Zhang et~al.(2021)Zhang, Guo, and Kordjamshidi}]{zhang2021towards}
Yue Zhang, Quan Guo, and Parisa Kordjamshidi. 2021.
\newblock Towards navigation by reasoning over spatial configurations.
\newblock In \emph{Proceedings of Second International Combined Workshop on Spatial Language Understanding and Grounded Communication for Robotics}, pages 42--52.

\bibitem[{Zhang et~al.(2024{\natexlab{c}})Zhang, Guo, and Kordjamshidi}]{zhang2024navhint}
Yue Zhang, Quan Guo, and Parisa Kordjamshidi. 2024{\natexlab{c}}.
\newblock Navhint: Vision and language navigation agent with a hint generator.
\newblock In \emph{Findings of the Association for Computational Linguistics: EACL 2024}, pages 92--103.

\bibitem[{Zhang and Kordjamshidi(2022{\natexlab{a}})}]{zhang2022explicit}
Yue Zhang and Parisa Kordjamshidi. 2022{\natexlab{a}}.
\newblock Explicit object relation alignment for vision and language navigation.
\newblock In \emph{Proceedings of the 60th Annual Meeting of the Association for Computational Linguistics: Student Research Workshop}, pages 322--331.

\bibitem[{Zhang and Kordjamshidi(2022{\natexlab{b}})}]{zhang2022lovis}
Yue Zhang and Parisa Kordjamshidi. 2022{\natexlab{b}}.
\newblock Lovis: Learning orientation and visual signals for vision and language navigation.
\newblock In \emph{Proceedings of the 29th International Conference on Computational Linguistics}, pages 5745--5754.

\bibitem[{Zhang and Kordjamshidi(2023)}]{zhang2023vln}
Yue Zhang and Parisa Kordjamshidi. 2023.
\newblock Vln-trans: Translator for the vision and language navigation agent.
\newblock In \emph{Proceedings of the 61st Annual Meeting of the Association for Computational Linguistics (Volume 1: Long Papers)}, pages 13219--13233.

\bibitem[{Zhang et~al.(2024{\natexlab{d}})Zhang, Ma, Li, Qiao, Wang, Chai, Wu, Bansal, and Kordjamshidi}]{zhang2024visionandlanguage}
Yue Zhang, Ziqiao Ma, Jialu Li, Yanyuan Qiao, Zun Wang, Joyce Chai, Qi~Wu, Mohit Bansal, and Parisa Kordjamshidi. 2024{\natexlab{d}}.
\newblock Vision-and-language navigation today and tomorrow: A survey in the era of foundation models.
\newblock \emph{Transactions on Machine Learning Research}.
\newblock Survey Certification.

\bibitem[{Zhang et~al.()Zhang, Xu, Shen, Kordjamshidi, and Huang}]{zhangspartun3d}
Yue Zhang, Zhiyang Xu, Ying Shen, Parisa Kordjamshidi, and Lifu Huang.
\newblock Spartun3d: Situated spatial understanding of 3d world in large language model.
\newblock In \emph{The Thirteenth International Conference on Learning Representations}.

\bibitem[{Zheng et~al.(2024)Zheng, Huang, Zhao, Zhong, and Wang}]{zheng2024towards}
Duo Zheng, Shijia Huang, Lin Zhao, Yiwu Zhong, and Liwei Wang. 2024.
\newblock Towards learning a generalist model for embodied navigation.
\newblock In \emph{Proceedings of the IEEE/CVF Conference on Computer Vision and Pattern Recognition}, pages 13624--13634.

\bibitem[{Zhou et~al.(2024{\natexlab{a}})Zhou, Hong, Wang, Wang, and Wu}]{zhou2024navgpt2}
Gengze Zhou, Yicong Hong, Zun Wang, Xin~Eric Wang, and Qi~Wu. 2024{\natexlab{a}}.
\newblock Navgpt-2: Unleashing navigational reasoning capability for large vision-language models.
\newblock In \emph{European Conference on Computer Vision}, pages 260--278. Springer.

\bibitem[{Zhou et~al.(2024{\natexlab{b}})Zhou, Hong, and Wu}]{zhou2024navgpt}
Gengze Zhou, Yicong Hong, and Qi~Wu. 2024{\natexlab{b}}.
\newblock Navgpt: Explicit reasoning in vision-and-language navigation with large language models.
\newblock In \emph{Proceedings of the AAAI Conference on Artificial Intelligence}, volume~38, pages 7641--7649.

\bibitem[{Zhou et~al.(2024{\natexlab{c}})Zhou, Hong, and Wu}]{zhou2023navgpt}
Gengze Zhou, Yicong Hong, and Qi~Wu. 2024{\natexlab{c}}.
\newblock \href {https://doi.org/10.1609/AAAI.V38I7.28597} {Navgpt: Explicit reasoning in vision-and-language navigation with large language models}.
\newblock pages 7641--7649. {AAAI} Press.

\end{thebibliography}
